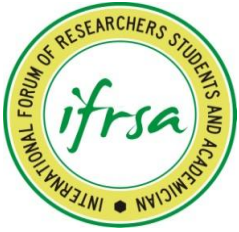

# IJGIP

# Content Based Image Indexing and Retrieval

Avinash N Bhute[1], B. B. Meshram[2]
[1,2]VJTI, Matunga, Mumbai

**ABSTRACT**

In this paper, we present the efficient content based image retrieval systems which employ the color, texture and shape information of images to facilitate the retrieval process. For efficient feature extraction, we extract the color, texture and shape feature of images automatically using edge detection which is widely used in signal processing and image compression. For facilitated the speedy retrieval we are implements the antipole-tree algorithm for indexing the images.

## 1. INTRODUCTION

As in internet era most difficult task is to retrieve the relevant information in response to a query. To help a user in this context various search system/engine are there in market with different features. In web search era 1.0 the main focus was on text retrieval using link analysis. It was totally read only era. There was no interaction in between the user and the search engine i.e. after obtaining search result user have no option to provide feedback regarding whether the result is relevant or not. In web search era 2.0 the focus was on retrieval of data based on relevance ranking as well as on social networking to read, write, edit and publish the result. Due to Proliferation of technology the current search era based on contextual search. Where rather than ranking of a page focus is on content based similarity to provide accurate result to user.

Content-based image retrieval (CBIR), also known as query by image content (QBIC) and content-based visual information retrieval (CBVIR) is the application of computer vision to the image retrieval problem, that is, the problem of searching for digital image in large databases. "Content-based" means that the search will analyse the actual contents of the image. The term 'content' in this context might refer to colors, shapes, textures, or any other information that can be derived from the image itself. Without the ability to examine image content, searches must rely on metadata such as captions or keywords, which may be laborious or expensive to produce.

In the past decade, many image retrieval systems have been developed, such as the IBM QBIC system developed at the IBM Almaden Research Center, the VIRAGE System developed by the Virage Incorporation, the Photobook System developed by the MIT Media Lab, the VisualSeek system developed at Columbia University, the WBIIS System developed at Stanford University, and the Blobworld System developed at U.C. Berkeley. With the expansion of digital image use, many researchers have been investigating increase in the efficiency of searching and indexing image data. Traditional text-based retrieval is not adequate for visual data. The common ground for them is to extract a signature for every image based on its pixel values, and to define a rule for comparing them. The signature can be shape, texture, color or any other information with which two images could be compared.

Approaches to content-based retrieval have taken two directions. In the first, image contents are modeled as a set of attributes extracted manually and managed within the framework of conventional database-management systems. Queries are specified using these attributes. Attribute-based representation of images entails a high level of image abstraction. Generally, the higher the level of abstraction, the lesser is the scope for posing ad hoc queries to the image database. Attribute-based retrieval is advocated and advanced primarily by database researchers. The second approach depends on an integrated feature-extraction/object-recognition subsystem to overcome the limitations of attribute-based retrieval. This subsystem automates the feature-extraction and object-recognition task that occurs when the image is inserted into the database. However, automated approaches to object recognition are computationally expensive, difficult, and tend to be domain specific. This approach is advanced primarily by image-interpretation researchers.

More effective techniques than simple browsing are necessary for searching collections of large numbers of images. An initial approach for organizing such image collections is to use words that refer to properties of the image, such as the creator, the place, the time, or the subject of the image.
Thetechniquethatisbasedonwordsforimageretrievaliscalledtext-basedimageretrievalormetadata-basedimageretrieval and constitutes a traditional technique that has been used during previous times for analog image collections.

The technique for imager etrieval from a digital collection by usingfeature – element values that are extracted automatically from the optical contents of the images is called content-based image retrieval. Feature extraction and analysis is performed from the images so that resulting values are comparable by the use of a computing machine for examining the similarity between images. Useful features for content-based image retrieval are considered those that mimic the features seen by humans, those that are perceived by the human vision. The use of such optical features, that reflect a view of image similarity as this is perceived by a man, even if he has difficulty in describing these features, increases the probability that the system recalls images that are similar, or alike, according to the human perception.

The features that are used for content-based image retrieval recharacterized as global (local) when they refer to the whole (apart of the) image. The basic characteristics that are used for content-based image retrieval are: the color (the distribution, or analogy of different colors at parts, or the whole image), the shape (the shape of the boundaries, or the interiors of objects depicted in the image), the texture (the presence of visual patterns that have properties of homogeneity and do not result from the presence of single color, or intensity), the location (the relative to other objects, or absolute position where each object resides in the image).

The key role in CBIR (Content Based Image Retrieval) system is image segmentation. The two most common strategies for image segmentation are color based and texture based. The color based search suffers a drawback that it clusters nearby elements as one part and usually ends up clustering two close by, but different objects. Texture based strategy on the other hand suffers poor segmentation. General approach towards CBIR is as shown in figure 1.The rest of the paper is organized into the following three sections. Rest of the paper is organized as follows- section two deals with the feature extraction algorithms, in section three we describe shape based retrieval, section presents similarity measures, in section five we discuss indexing and retrieval and finally we conclude in section six.

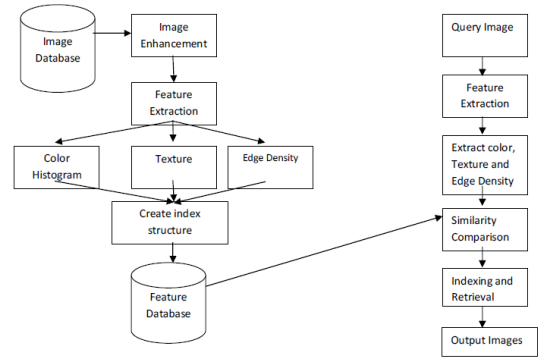

**Fig.1. General approaches towards CBIR**

As shown in figure 1 from the image database image features like color, texture, edge density, etc. are extracted after preprocessing image data which may include image enhancement, noise removal etc. These features are then stored as indexed feature database to enable efficient image retrieval. Same features are also extracted from user submitted query and similarity measures are employed to find similarity between two images. Best matched ranked images from image database are presented to the user.

The popular algorithms may be used for implementation various stages are tabulated in table 1.

| Processing at Database Server | |
|---|---|
| Features | Popular Algorithm used |
| Color | GCH,Color correlogram |
| Texture | Homogeneity, Entropy Energy, Harr Wavelet |
| Edge | Sobel operation |
| Shape | Eccentricity |
| Index structure creation | Antipole tree |
| Processing for Input | |
| Similarity matching | Euclidean distance |
| Indexing and Retrieval | Antipole tree and Range Search |

**Table 1,. CBIR Features and Retrieval algorithm**

## 2. FEATURE EXTRACTION ALGORITHMS

In this section we discuss various features that can be extracted from digital images for indexing and retrieval.

*A. Color Feature*

A digital image may be considered as a two dimensional array where the array cells correspond to the image pixels and the values stored in the cells to the values of color-intensity, in case of a grey scale (single-color) image[2]. A color image consists of three single-color images that correspond to the colors Red, Green and Blue (from which any color may be composed, when appropriate intensity values are combined). By making a function from the discrete values of intensity to the number of pixels with the respective value, we construct a Histogram for each of the component colors [14].To achieve low computational complexity several neighbouring values of intensities can be grouped together to reduce number of histogram bins.

### i. Image Histogram

An image histogram *refers* to the probability mass function of the image intensities. This is extended for color images to capture the joint probabilities of the intensities of the three color channels. More formally, the color histogram *h* is defined by,

$$h_{A,B,C}(a,b,c) = N \cdot \Pr ob(A = a, B = b, C = c)$$

Where *A, B* and *C* represent the three color channels (R,G,B or H,S,V) and *N* is the number of pixels in the image. Computationally, the color histogram is formed by discretizing the colors within an image and counting the number of pixels of each color. Since the typical computer represents color images with up to 224 colors, this process generally requires substantial quantization of the color space. The main issues regarding the use of color histograms for indexing involve the choice of color space and quantization of the color space. When a perceptually uniform color space is chosen uniform quantization may be appropriate. If a non-uniform color space is chosen, then non-uniform quantization may be needed[5].

**Color Histogram Generation Algorithm**
Color Histogram Generation
The color histogram for an image is constructed by counting the number of pixels of each color.
Input: Query Image
Output: Color Histogram of an image

1. Scale the image to size 512 x 512 to normalize the histogram.
2. Set up parameters for the Histogram object.
3. bin← number of bins in histogram.
4. low← lowest value of the bin.
5. high←highest value of bin.
6. Create a object of class histogram
7. Create a Histogram operation with the required parameters or create a ParameterBlock with the parameters and pass it to the Histogram operation.
8. The histogram data stored in the object.
9. The data consists of:
10. Number of bands in the histogram
11. Number of bins for each band of the image
12. Lowest value checked for each band
13. Highest value checked for each band

Pseudo code to compare Two Histograms
Input: two objects of type Histogram.
Output: Euclidian distance between two histograms.
Histogram Comparison
  dist: Euclidian distance
  Let sum=0
  Until temp<256 repeat step 4
  sum = sum + square(hist1(temp)-hist2(temp))
  dist=sqrt(sum)
  returndist;

The color histogram can be thought of as a set of vectors. For grey-scale images these are two dimensional vectors. One dimension gives the value of the grey-level and the other the count of pixels at the grey-level. For color images the color histograms are composed of 4-D vectors. This makes color histograms very difficult to visualize. There are several lossy approaches for viewing color histograms; one of the easiest is to view separately the histograms of the color channels.The main limitation of color histogram approach is its inability to distinguish images whose color distribution is identical but whose pixels are organized according to a different layout

### ii. Color Correlogram

The Color Correlogram of an image is the probability of joint occurrence of two pixels some distance apart that one pixel belongs to a specific color and the other belongs to same or another color[7].The color correlogram, a feature originally proposed by Huang, overcomes this limitation by encoding color-spatial information into co-occurrence matrices. Each entry (i, j) in the co-occurrence matrix expresses how many pixels whose color is $C_j$ can be found at a distance d from a pixel whose color is $C_i$. Each different value of d leads to different co-occurrence matrix.

### B. Texture Feature

Texture is another important property of images. Various texture representations have been investigated in pattern recognition and computer vision. Texture representation methods can be classified into two categories: structural and statistical. Structural methods, including morphological operator and adjacency graph, describe texture by identifying structural primitives and their placement rules. They tend to be most effective when applied to textures that are very regular. Statistical methods, including Fourier power spectra, co-occurrence matrices, shift-invariant principal component analysis (SPCA), Tamura feature, Wolds decomposition, Markov random field, fractal model, and multi-resolution filtering techniques such as Gabor and Haar wavelet transform, characterize texture by the statistical distribution of the image intensity[23]. The Extraction of feature vector is the most crucial step in the whole CBIR system. This is because these feature vectors are used in all the subsequent modules of the system. It is to be realized that the image itself plays no part in the following modules. It is the feature vectors that are dealt with. The quality of the output drastically improves as the feature vectors that are used are made more effective in representing the image. Texture of image provides following features of the image.

Energy = $\sum_i \sum_j P^2_d(i,j)$

Entropy = $-\sum_i \sum_j P_d(i,j) \log P_d(i,j)$
Contrast = $\sum_i \sum_j (i-j)^2 P_d(i,j)$
Homogeneity = $\sum_i \sum_j P_d(i,j) / 1+ |i-j|$
Where P is Co-occurrence Matrix.

### i. Haar Wavelets

The Wavelets are useful for hierarchically decomposing functions in ways that are both efficient and theoretically sound. Broadly speaking, a wavelet representation of a function consists of a coarse overall approximation together with detail coefficients that influence the function at various scaled. The wavelet transform has excellent energy compaction and de-correlation properties, which can be used to effectively generate compact representations that exploit the structure of data[12]. By using wavelet sub band decomposition, and storing only the most important sub bands (that is, the top coefficients), we can compute fixed-size low-dimensional feature vectors independent of resolution, image size and dithering effects. In addition, wavelets are robust with respect to color intensity shifts, and can capture both texture and shape information efficiently. Furthermore, wavelet transforms can be computed in linear time, thus allowing for very fast algorithms. Haar wavelets enable speed up the wavelet computation phase for thousands of sliding windows of varying sizes in an image. They also facilitate the development of efficient incremental algorithms for computing wavelet transforms for larger windows in terms of the ones for smaller windows. One disadvantage of Haar wavelets is that it tends to produce large number of signatures for all windows in image. Figure 2 shows the result of haar transformation on any image.

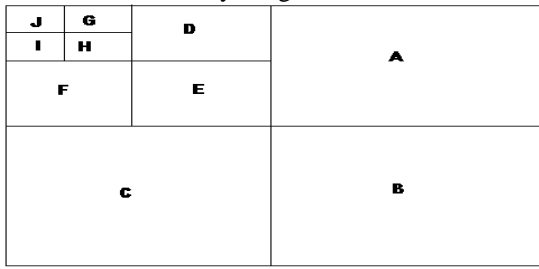

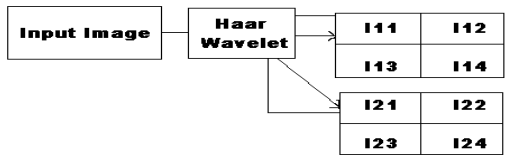

**Figure 2. HAAR Wavelet and its transformation**

In each iteration Ii2...4 images are saved and Ii1 sub image is again subjected to wavelet Transformation instead of entire image for three iterations, by which 10 sub images of input image are obtained. Sub image I11 is further divided into sub images I21 ... I24 in the second iteration. The sub image I21 is further divided into I31 I32 I33 I34 in the third iteration. All sub images are normalized to maintain the uniform size.

**Wavelet Signature Generation Algorithm**
Input: Image whose Haar transform has to be calculated.
Output: Wavelet signature of the image

**Wavelet Signature Genetation**
    Let I→ the input image
    w x w→ size of image
    Let matrix[i][j] contain intensity of pixel at position i , j.
    Call haar2 function to perform 2 dimensional Haar transform.
    Let I1, I2, I3, I4 based on Haar wavelet of size w/2×w/2
    Compute Signatures for I2, I3, I4.
Now take the image I1 and divide it into 4 bands namely I11, I12, I13, I14 of size w/4×w/4
Compute signatures for I12, I13, I14.
Again take the I11 and divide it into 4 bands namely I111, I112, I113, I114 of size w/8×w/8.
Now we obtain 10 signatures then stop the process.

**Algorithm to calculate wavelet signature**
Input: Image I
Output: Wavelet Signature "sign"

Calculation of wavelet signature
    Let w= width of image.
    h= height of image.
    sum=0, i=0, j=0
    until i<w repeat till step 7
    until j<h repeat till step 6
    sum=sum + intensity at (i , j)
    j = j+1
    i = i+1
    sign=sqrt(sum)
    return sign
Where "sign" is the computed Wavelet signature (texture feature representation) of the sub image.
In the following figure 3, Haar transform is performed on image and its result is shown.
    EXAMPLE:

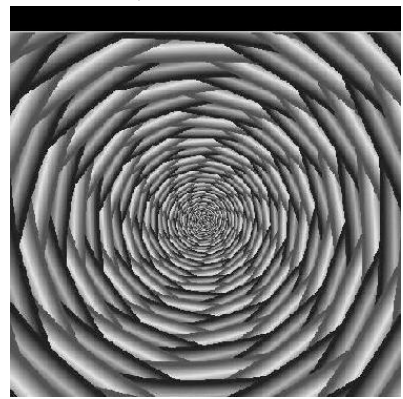

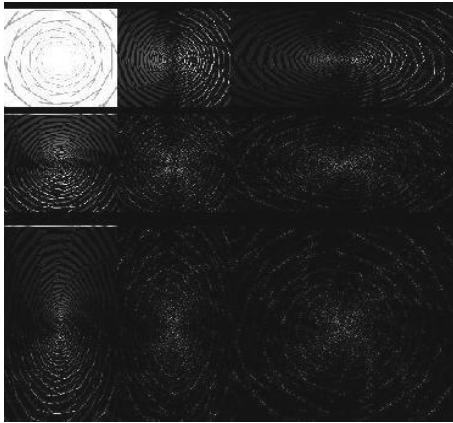

**Figure 4. Input image for Haar wavelet and its result**

In Texture Based Medical Image Indexing and Retrieval [By TristanGlatard, Johan Montagnat, Isabelle E. Magnin], authors analyze medical image properties and evaluate Gabor-filter based features extraction for medical images indexing and classification. The goal is to perform clinically relevant queries on large image databases that do not require user supervision.They also demonstrates on the concrete case of cardiac imaging that these techniques can be used for indexing, retrieval by similarity queries, and to some extent, extracting clinically relevant information out of the images.

### C. Edge Feature

Edge detection is useful for locating the boundaries of objects within an image. Any abrupt change in image frequency over a relatively small area within an image is defined as an edge[6]. Image edges usually occur at the boundaries of objects within an image, where the amplitude of the object abruptly changes to the amplitude of the background or another object. The GradientMagnitude operation is an edge detector that computes the magnitude of the image gradient vector in two orthogonal directions. This operation is used to improve an image by showing the directional information only for those pixels that have a strong magnitude for the brightness gradient.

It performs two convolution operations on the source image. One convolution detects edges in one direction, the other convolution detects edges the orthogonal direction. These two convolutions yield two intermediate images.

It squares all the pixel values in the two intermediate images, yielding two more intermediate images.

It takes the square root of the last two images forming the final image.

Edge Detection using Gaussian filtering

In order to perform edge detection on an image Gaussian filter followed by Sobel operation is performed.

Gaussian Filtering

The Gaussian blur is a type of image-blurring filter that uses a normal distribution (also called "Gaussian distribution", thus the name "Gaussian blur") for calculating the transformation to apply to each pixel in the image[26]. The equation of Gaussian distribution in N dimensions is

$$G(r) = \frac{1}{(2\pi\sigma^2)^{N/2}} e^{-r^2/(2\sigma^2)}$$

Or specifically in two dimensions

$$G(u,v) = \frac{1}{2\pi\sigma^2} e^{-(u^2+v^2)/(2\sigma^2)}$$

Where, $r$ is the blur radius ($r^2 = u^2 + v^2$)
$\sigma$ is the standard deviation

Step 1.1: Values from this distribution are used to build a convolution matrix which is applied to the original image. Our convolution matrix for a 3x3 mask will be

$$\frac{1}{16} \begin{vmatrix} 1 & 2 & 1 \\ 2 & 4 & 2 \\ 1 & 2 & 1 \end{vmatrix}$$

Step 1.2: Each pixel's new value is set to a weighted average of that pixel's neighbourhood. The original pixel's value receives the heaviest weight (having the highest Gaussian value) and neighbouring pixel receives smaller weights as their distance to the original pixel increases. This results in a blur that preserves boundaries and edges. Consider a sample image given below

| 54 | 46 | 55 | 54 | 46 |
|---|---|---|---|---|
| 22 | 22 | 22 | 22 | 22 |
| 100 | 100 | 100 | 100 | 100 |
| 120 | 125 | 125 | 125 | 125 |
| 125 | 125 | 125 | 125 | 125 |

Applying the above 3x3 mask to the image
Calculation 1/16

| (54*1) | (46*2) | (55*1) |
|---|---|---|
| (22*2) | (22*4) | (22*2) |
| (100*1) | (100*2) | (100*1) |

| 54 | 46 | 55 | 54 | 46 |
|---|---|---|---|---|
| 22 | 22 | 22 | 22 | 22 |
| 100 | 100 | 100 | 100 | 100 |
| 120 | 125 | 125 | 125 | 125 |
| 125 | 125 | 125 | 125 | 125 |

The resultant matrix would be

= 1/16 

| 54 | 92 | 55 |
|---|---|---|
| 44 | 88 | 44 |
| 100 | 200 | 100 |

= 1/16(54+92+55+44+88+44+100+200+100)= 49

Therefore 22 will be replaced by 49

Similarly the mask will move across the whole matrix and the resultant output image will be

| 54 | 46 | 55 | 54 | 46 |
|---|---|---|---|---|
| 49 | 49 | 49 | 49 | 49 |
| 86 | 86 | 86 | 86 | 86 |
| 118 | 118 | 119 | 119 | 119 |
| 125 | 125 | 125 | 125 | 125 |

*i. Sobel's Operator*

After smoothing the image and eliminating the noise, the next step is to find the edge strength by taking the gradient of the image. The Sobel operator performs a 2-D spatial gradient measurement on an image. Then, the approximate absolute gradient magnitude (edge strength) at each point can be found.

Step 2.1

The Sobel operator uses a pair of 3x3 convolution masks, one estimating the gradient in the x-direction (columns) and the other estimating the gradient in the y-direction (rows). They are shown below

| -1 | 0 | +1 |
|---|---|---|
| -2 | 0 | +2 |
| -1 | 0 | +1 |

| +1 | +2 | +1 |
|---|---|---|
| 0 | 0 | 0 |
| -1 | -2 | -1 |

Step 2.2 Now the magnitude is calculated using the following formulae

$G = \sqrt{(G_x^2 + G_y^2)}$

Example let our image matrix be

| 0 | 30 | 60 |
|---|---|---|
| 5 | 32 | 62 |
| 10 | 38 | 64 |

Applying Sobel's mask to the matrix at 32

| 0 | 30 | 60 |
|---|---|---|
| 5 | <u>32</u> | 62 |
| 10 | 38 | 64 |

Calculation of Gx at 32

| (-1*0) | (0*30) | (1*60) |
|---|---|---|
| (-2*5) | (0*32) | (2*62) |
| (-1*10) | (0*38) | (1*64) |

=(0+0+60-10+0+124-10+0+64)=228

Calculation of Gy at 32

| (-1*0) | (-2*30) | (-1*60) |
|---|---|---|
| (0*5) | (0*32) | (0*62) |
| (1*10) | (2*38) | (1*64) |

=(0-60-60+0+0+0+10+76+64)=30

Similarly the mask will move across the whole matrix and Gx and Gy will be       GxGy

| 117 | 237 | 120 |
|---|---|---|
| 112 | 228 | 146 |
| 111 | 219 | 108 |

| 17 | 11 | 8 |
|---|---|---|
| 38 | 30 | 20 |
| 21 | 19 | 8 |

Now the magnitude is calculated using the formulae
$G = \sqrt{(G_x^2 + G_y^2)}$

$117^2 + 17^2 = 118$   $237^2 + 11^2 = 237$   $120^2 + 8^2 = 120$
$112^2 + 38^2 = 118$   $228^2 + 30^2 = 230$   $146^2 + 20^2 = 147$
$111^2 + 21^2 = 113$   $219^2 + 19^2 = 220$   $108^2 + 8^2 = 108$

Thus, we have the magnitude or gradient strength of each pixel.

**Algorithm to perform Sobel Operation**
Input: Input image
Output: Set of points on edge
**Sobel Masking**
1. Scale the image to size 100 x 100.
2. Convert the image to gray scale image
3. Create three kernal

a. floatdata_g[] = newfloat[] { 0.625F, 0.125F, 0.625F,0.125F, 0.25F, 0.125F,0.625F, 0.125F, 0.625F};
b. floatdata_h[] = newfloat[] { 1.0F, 0.0F, -1.0F, 1.414F, 0.0F, -1.414F,1.0F, 0.0F, -1.0F};
c. floatdata_v[] = newfloat[] {-1.0F, -1.414F, -1.0F, 0.0F, 0.0F, 0.0F,1.0F, 1.414F, 1.0F};
KernelJAIkern_g = newKernelJAI(3,3,data_g);
KernelJAIkern_h = newKernelJAI(3,3,data_h);
KernelJAIkern_v = newKernelJAI(3,3,data_v);

Apply kernel on input image
5. Let i=0, j=0
6. until i<100 repeat till step 10
7. until j<100 repeat till step 11
8. if intensity(i ,j)==255
9. add (i , j) to vector V
10. j = j + 1
11. i = i + 1

### 3. SHAPE-BASED IMAGE RETRIEVAL

Shape extraction remains a challenge to feature-oriented approaches. Several methods have been developed for detecting shapes indirectly. Whereas it tends to be extremely difficult to perform semantically meaningful segmentation, many reasonably reliable algorithms for low-level feature extraction have been developed. These will be used to provide the spatial information that is lacking in color histograms.

Rather than attempt to directly measure shape we calculate some simpler properties that are indirectly related to shape and avoid the requirement for good segmentation, providing a more practical solution.

Edge Orientation

We may combine edge orientation histograms with color histograms. These edge orientation histograms encode some aspects of shape information. As a result, image retrieval can be more responsive to the shape content of the images. Standard edge detection is sufficient for shape-oriented feature extraction. In addition, minor errors in the edge map have little effect on the edge orientation histograms. Unlike color histograms, the orientation histograms are not rotationally invariant. Therefore the histogram matching process has to iteratively shift the histogram to find the best match[6].

A more important consideration is that the edge maps were thresholded by some unspecified means. For robustness an adaptive thresholding scheme should be used. However, an alternative is to include all the edges and weight their contribution to the histogram by their magnitudes so as to reduce the contribution from spurious edges. This is the approach we take in the reported experiments.

Multi-resolution Salience Distance Transform

Another approach to including shape information is based on the distance transform (DT). The DT is a method for taking a binary image of feature and non-feature pixels and calculating at every pixel in the image the distance to the closest feature. Although this is a potentially expensive operation efficient algorithms have been developed that only require two passes through the image.

To improve the stability of the distance transform, Rosin and West developed an algorithm called the salience distance transform (SDT). In SDT, the distances are weighted by the salience of the edge, rather than propagating out Euclidean (or quasi-Euclidean) distances from edges. Various forms of salience have been demonstrated, incorporating features such as edge magnitude, curve length, and local curvature. The effect of including salience was to downplay the effect of spurious edges by soft assignment while avoiding the sensitivity problems of thresholding.

The distance values can be represented in histograms once the SDT has been performed. These histograms will respond differently to different types of shapes. There is the crude distinction between cluttered, complex scenes and simple sparse scenes, which will result in different ends of the histogram being heavily populated. Thus the profile of the distance histograms provides an indication of image complexity. However, rather than return a single complexity measurement, the shape of the histogram will indicate more subtle distinctions between shapes.

### 4. SIMILARITY MEASUREMENT

Once the features are extracted and stored and or indexed for retrieval, we need to find distance between database image and user query input image. There are various approaches for this. In this section we describe these approaches.

*A. Histogram Euclidean distance*

Let *h* and *g* represent two color histograms. The Euclidean distance between the color histograms *h* and *g* can be computed as:

$$d^2(h,g) = \sum_A \sum_B \sum_C (h(a,b,c) - g(a,b,c))^2$$

In this distance formula, there is only comparison between the identical bins in the respective histograms. Two different bins may represent perceptually similar colors but are not compared cross-wise. All bins contribute equally to the distance[19].

*B. Histogram Intersection Distance*

The color histogram intersection was proposed for color image retrieval. The intersection of histograms *h* and *g* is given by:

$$d(h,g) = \frac{\sum_A \sum_B \sum_C \min(h(a,b,c), g(a,b,c))}{\min(|h|,|g|)} \quad (5)$$

where |h| and |g| gives the magnitude of each histogram, which is equal to the number of samples. Colors not present in the user's query image do not contribute to the intersection distance. This reduces the contribution of background colors. The sum is normalized by the histogram with fewest samples[30].

### C. Housdorff distance

Named after Felix Housdorff (1868-1942), Housdorff distance is the « *maximum distance of a set to the nearest point in the other set* ]. More formally, Housdorff distance from set A to set B is a *maximin* function, defined as

$$h(A,B) = \max_{a \in A} \{ \min_{b \in B} \{ d(a,b) \} \} \quad (eq.\ 2)$$

where a and b, are points of sets A and B respectively, and d(a, b) is any metric between these points ; for simplicity, we'll take d(a, b) as the Euclidian distance between a and b. If for instance A and B are two sets of points

## 5. IMAGE INDEXING AND SEARCHING

This section describes the most commonly used antipole tree and range search algorithm for indexing and searching.

Range query: Given a query object q, a database S, and a threshold t, the Range Search problem is to find all objects {o € S |dist(o,q)<t} [8]

k-Nearest Neighbor query: Given a query object q and an integer k > 0, the k-Nearest Neighbor Problem is to retrieve the k closest elements to q in S [8].

### A. THE ANTIPOLE TREE

Let (M, dist) be a finite metric space, let S be a subset of M, and suppose that we aim to split it into the minimum possible number of clusters whose radii should not exceed a given threshold. The Antipole clustering of bounded radius is performed by a recursive top-down procedure starting from the given finite set of points S and checking at each step if a given splitting condition Φ is satisfied. If this is not the case, then splitting is not performed, the given subset is a cluster, and centroid having distance approximately less than from every other node in the cluster is computed. Otherwise, if is satisfied then a pair of points {A,B} of S, called the Antipole pair, is generated as explained in the following section and is used to split S into two subsets $S_A$ and $S_B$ obtained by assigning each point p of S to the subset containing the endpoint closest to p of the Antipole (A;B). The splitting condition Φ states that dist(A,B) is greater than the cluster diameter threshold corrected by an error. The tree obtained by the above procedure is called an Antipole Tree. All nodes are annotated with the Antipole endpoints and the corresponding cluster radius; each leaf contains also the 1-median of the corresponding final cluster.

### B. 1-MEDIAN CALCULATION

Following is the algorithm for the approximate 1-median selection, an important subroutine in our Antipole Tree construction. It is based on a tournament played among the elements of the input set S. At each round, the elements which passed the preceding turn are randomly partitioned into subsets, say $X_1,…,X_k$. Then, each subset $X_i$ is locally processed by a procedure which computes its exact 1-median $x_i$. The elements $x_1, . . . ,x_k$ move to the next round. The tournament terminates when we are left with a single element x, the final winner. The winner approximates the exact 1-median in S. The local optimization procedure 1-MEDIAN (X) returns the exact 1-median in X. A running time analysis shows that the above procedure takes time *n + o(n) in the worst case*.

ALGORITHM
Purpose: To find 1- median of elements of given input set S.
    Input: Set of elements S.
    Output: 1-median of set S.

    1-Median Calculation
1-MEDIAN(X)
1. For each x € X do
2. $S_x \leftarrow \sum_{y \in x} dist(x,y)$;
3. Let m € X be an element such that $S_m$ = min $_{x \in X}(S_x)$;
4. return m;
5. END 1- MEDIAN;

Algorithm 2
APPROX_1_MEDIAN(S)
1. while |S| > threshold do
2. W←Ø;
3. while |S| 2τ do
4. choose randomly a subset T S, with |T|= τ
5. S←S\T;
6. W←W {1- MEDIAN(T)};
7. End while;
8. return 1-MEDIAN(S);
9. END APPROX_1_MEDIAN

### C. THE DIAMETER (ANTIPOLE) COMPUTATION

Let (M, d) be a metric space with distance function dist: (M X M)→IR and let S be a finite subset of M. The diameter computation problem or furthest pair problem is to find the pair of points A, B in S such that dist(A,B) ≥ dist(x, y) x; y S.

We can construct a metric space where all distances among objects are set to 1 except for one (randomly chosen) which is set to 2. In this case, any algorithm that tries to give an approximation factor greater than 1/2 must examine all pairs, so a randomized algorithm will not necessarily find that pair.

Nevertheless, we expect a good outcome in nearly all cases. Here, we introduce a randomized algorithm inspired by the one proposed for the 1-median computation and reviewed in the preceding section. In this case, each subset $X_i$ is locally processed by a procedure LOCAL WINNER which computes its exact 1-median $x_i$ and then returns the set $X_i$, obtained by removing the element $x_i$ from $X_i$. The elements in $X_{1X2}$ . . . $X_k$ are used in the subsequent step. The tournament terminates when we are left with a single set, X, from which we extract the final winners A, B, as the furthest points in X. The pair A, B is called the Antipole pair and their distance represents the approximate diameter of the set S.

The pseudo code of the Antipole algorithm APPROX ANTIPOLE similar to that of the 1-Median algorithm is given in below. A faster (but less accurate) variant of APPROX ANTIPOLE can be used. Such variant, called FAST APPROX ANTIPOLE consists of taking $X_i$ as the farthest pair of $X_i$. Its pseudo code can therefore be obtained simply by replacing in APPROX ANTIPOLE each call to LOCAL WINNER by a call to FIND ANTIPOLE. In the next section, we will prove that both variants have a linear running time in the number of elements. We will also show that FAST APPROX ANTIPOLE is also linear in the tournament size τ, whereas APPROX ANTIPOLE is quadratic with respect to τ.

For tournaments of size 3, both variants plainly coincide. Thus, since in the rest of the paper only tournaments of size 3 will be considered, by referring to the faster variant we will not lose any accuracy.

ALGORITHM
LOCAL_WINNER (T)
    1. return T \ 1-MEDIAN(T)
    2. END LOCAL_WINNER
FIND_ANTIPOLE (T)
    1. return $P_1$, $P_2$ T such that
        dist $(P_1,P_2)$ dist(x, y) x, y T
    2. END FIND_ANTIPOLE
APPROX_ANTIPOLE(S)
    1. while |S| > threshold do
    2. W←Ø;
    3. while |S| 2τ do
    4. choose randomly a subset T S, with |T|= τ
    5. S←S\T;
    6. W←W {LOCAL_WINNER(T)};
    7. End while;
    8. S←W {LOCAL_WINNER(S)};
    9. return FIND_ANTIPOLE(S);
    10. END APPROX_ANTIPOLE

### D. RANGE SEARCH ALGORITHM

The range search algorithm takes as input the Antipole Tree T, the query object q, the threshold t, and returns the result of the range search of the database with threshold t. The search algorithm recursively descends all branches of the tree until either it reaches a leaf representing a cluster to be visited or it detects a sub tree that is certainly out of range and, therefore, may be pruned out. Such branches are filtered by applying the triangle inequality. Notice that the triangle inequality is used both for exclusion and inclusion. The use for exclusion establishes that an object can be pruned, thus avoiding the computation of the distance between such an object and the query. The other usage establishes that an object must be inserted because the object is close to its cluster's centroid and the centroid is very close to the query object.

**Input**: Antipole Tree T, query object q, the threshold t.
**Output**: The result of the range search of the database with threshold t.

    **Range Search algorithm**
    RANGE_SEARCH(T, q, t, OUT)
    1. If(T.Leaf=FALSE) then
    2. $D_A$← dist(q, T, A);
    3. $D_B$← dist(q, T, B);
    4. If($D_B \leq t$) then
    5.     OUT ← OUT (T.A);
    6. End if;
    7. If($D_B \leq t$) then
    8. OUT ← OUT (T.B);
    9. End if;
    10. $q.D_V$ ← $q.D_V$ ($D_A$, $D_B$);
    11. if($D_A \leq t + T.Rad_A$) then
    12. RANGE_SEARCH(T.left, q, t, OUT);
    13. End if;
    14. if($D_B \leq t + T.Rad_B$) then
    15. RANGE_SEARCH(T.right, q, t, OUT);
    16. End if;
    17. $q.D_V$ ← $q.D_V$ {$D_A$, $D_B$};
    18. return;
    19. else // leaf case
    20. OUT ← OUT{(T.Cluster, q, t, OUT)};
    21. End if:
    22. END RANGE_SEARCH

Following table summarizes various CBIR techniques.

TABLE 2. COMPARISON OF CBIR APPROACHES

| Algorithms | Approach | Advantages | Limitation | Comments |
|---|---|---|---|---|
| Image Retrieval using Color features | This approached is regarding image retrieval based on both color features and spatial distribution of image. | The clusters of the images are determined by initially pre-processing the image database. | | All the algorithm retrieved the images with own principles and have the advantages and limitations. Here we analyse four algorithms and choose the best one among all and describe the pseudo code for the particular algorithm. Here we select Sequential Multiple Attribute Tree (SMAT) algorithm because This is able to rank the candidate image accordingly. Its approach is a multi-tier tree structure, where each layer corresponds to an indexing attribute. For example, the top layer can be based on color, the second is based on color percentage or size of the cluster, and the last is based on spatial property. While other algorithm specify such as Text, Colour Histogram and spatial distribution of image. |
| Image retrieval using Image Color Histogram | The color histogram is easy to compute and effective in characterizing both the global and local distribution of colors in an image. | 1. It is robust to translation and rotation about the view axis and 2. Changes only slowly with the scale | 1. A histogram with a large number of bins will increase the computational cost. 2. A histogram with a large number of bins will be inappropriate for building efficient indexes for image databases. | |
| Image retrieval using Textual description | Most existing Image Retrieval systems are text-based, but images frequently have little or no accompanying textual information. The solution historically has been to develop text-based ontology and classification schemes for image description | Text-based indexing methods have much strength including the ability to represent both general and specific instantiations of an object at varying levels of complexity. | 1. Automatically generating descriptive texts for a wide spectrum of images is not feasible, most text-based retrieval systems require manual annotation of images. | |

## 6. CONCLUSION

We have presented a comprehensive survey highlighting popular methods and algorithms for evaluation relevant to the young and exciting field of image retrieval. With ease of bandwidth, memory and computational power, we believe that the field will experience a paradigm shift in the future, with the focus being more on application-oriented, generating considerable impact in day-to-day life. With explosive growth of social media, image search options being provided by most search engine has contributed in quality and quantity of images being uploaded in recent times. Although search engines at present do not work on content of the image. Meanwhile we hope that thrust for development of novel technologies in the field of image retrieval will continue. The future of CBIR will surely depend on collective focus and progress on integrating multiple features covering all aspects of images.

*Avinash N Bhute, B. B. Meshram| Content Based Image Indexing and Retrieval*